\documentclass[sigconf]{acmart}

\AtBeginDocument{%
  \providecommand\BibTeX{{%
    \normalfont B\kern-0.5em{\scshape i\kern-0.25em b}\kern-0.8em\TeX}}}

\setcopyright{acmlicensed}
\copyrightyear{2018}
\acmYear{2018}
\acmDOI{XXXXXXX.XXXXXXX}

\acmConference[Conference acronym 'XX]{Make sure to enter the correct
  conference title from your rights confirmation emai}{June 03--05,
  2018}{Woodstock, NY}
\acmISBN{978-1-4503-XXXX-X/18/06}

\usepackage{multirow}
\usepackage{subfigure}
\usepackage{color, colortbl}
\usepackage{booktabs}
\usepackage{caption}
\usepackage{subcaption}

\newtheorem{definition}{Definition}

\newcommand{\hide}[1]{} 

\newcommand{\vpara}[1]{\vspace{0.05in}\noindent\textbf{#1 }}

\definecolor{lgray}{gray}{0.95}
\definecolor{orange}{RGB}{224, 108, 0}
\definecolor{blue}{RGB}{25, 117, 210}

\begin{document}

\title{Graph Pre-Training Models Are Strong Anomaly Detectors}

\author{Jiashun Cheng}
\affiliation{%
  \institution{HKUST}
  \city{Hong Kong SAR}
  \country{China}
}
\email{jchengak@connect.ust.hk}

\author{Zinan Zheng}
\affiliation{%
  \institution{HKUST (GZ)}
 \country{}
  \city{Guangzhou}
  \country{China}
}
\email{zzheng078@connect.hkust-gz.edu.cn}

\author{Yang Liu}
\affiliation{%
  \institution{HKUST}
  \city{Hong Kong SAR}
  \country{China}
}
\email{yliukj@connect.ust.hk}

\author{Jianheng Tang}
\affiliation{%
  \institution{HKUST}
  \city{Hong Kong SAR}
  \country{China}
}
\email{jtangbf@connect.ust.hk}

\author{Hongwei Wang}
\affiliation{%
  \institution{Tencent AI Lab Seattle}
  \city{Seattle}
  \country{United States}
}
\email{hongweiw@tencent.com}

\author{Yu Rong}
\affiliation{%
  \institution{DAMO Academy, Alibaba Group}
  \city{Hangzhou}
  \country{China}
}
\email{yu.rong@hotmail.com}

\author{Jia Li}
\authornote{Corresponding author.}
\affiliation{%
  \institution{HKUST (GZ), HKUST}
  \city{Guangzhou/Hong Kong SAR}
  \country{China}
}
\email{jialee@ust.hk}

\author{Fugee Tsung}
\affiliation{%
  \institution{HKUST (GZ), HKUST}
  \city{Guangzhou/Hong Kong SAR}
  \country{China}
}
\email{season@ust.hk}


\begin{abstract}
  Graph Anomaly Detection (GAD) is a challenging and practical research topic where Graph Neural Networks (GNNs) have recently shown promising results.
  The effectiveness of existing GNNs in GAD has been mainly attributed to the simultaneous learning of node representations and the classifier in an end-to-end manner.
  Meanwhile, graph pre-training, the two-stage learning paradigm such as DGI and GraphMAE, has shown potential in leveraging unlabeled graph data to enhance downstream tasks, yet its impact on GAD remains under-explored.
  In this work, we show that graph pre-training models are strong graph anomaly detectors. Specifically, we demonstrate that pre-training is highly competitive, markedly outperforming the state-of-the-art end-to-end training models when faced with limited supervision.
  To understand this phenomenon, we further uncover pre-training enhances the detection of distant, under-represented, unlabeled anomalies that go beyond 2-hop neighborhoods of known anomalies, shedding light on its superior performance against end-to-end models.
  Moreover, we extend our examination to the potential of pre-training in graph-level anomaly detection.
  We envision this work to stimulate a re-evaluation of pre-training's role in GAD and offer valuable insights for future research.
\end{abstract}





\maketitle

\section{Introduction}

Anomalies, also known as outliers, exceptions, or novelties, represent a small but significant portion of data that deviates markedly from the standard, normal, or prevalent patterns \cite{textbook, anomaly_survey}. 
The identification of these anomalies is especially potent in the context of graphs, which depict relationships or interactions across various domains.
Graphs offer a multi-dimensional perspective, uncovering hidden connections and correlations that might be missed in the traditional independent and identically distributed data analysis. 

Traditional graph anomaly detection (GAD) methodologies, dependent on manually crafted features \cite{noble2003graph,akoglu2010oddball} and statistical models \cite{heard2010bayesian,aggarwal2013outlier}, struggled with identifying unknown anomalies and were often labor-intensive.
To address these challenges, there has been a shift towards advanced techniques like deep graph representation learning and Graph Neural Networks (GNNs). 
These approaches excel in distinguishing anomalies from normal data and autonomously learn deviating patterns through the extraction of rich, expressive representations from the graph. Moreover, several adaptations of standard GNNs have been proposed to address several unique challenges in GAD, such as label imbalance \cite{PCGNN}, feature heterophily \cite{BWGNN}, and relation camouflage \cite{CareGNN}.

On the other front, pre-training has emerged as a promising technique in the rapidly evolving domain of graph learning. 
This approach typically follows a two-step pipeline, starting with the acquisition of universal knowledge from unlabeled data through self-supervised objectives, and then applying this knowledge to specific downstream tasks. 
Two prominent pre-training approaches, contrastive learning~\cite{velivckovic2018deep, hassani2020contrastive, thakoor2022bgrl} and predictive learning~\cite{hou2022graphmae, cheng2023wiener, tan2023s2gae}, have shown great potential in facilitating more competitive and robust classification.

Nonetheless, in the field of GAD, there is a notable imbalance in the research focus. 
While numerous studies explore the learning architectures of GAD, investigations into learning paradigms, especially pretraining, are less common. 
For instance, a recent benchmark study \cite{tang2023gadbench} examined 29 different GAD models, but only one of these \cite{wang2021decoupling} incorporated pre-training. 
This indicates a significant gap in the thorough and systematic exploration of pre-training's effectiveness and impact on GAD. 
Two essential questions remain under-explored: 
(1) the concrete benefits of pre-training compared to end-to-end models trained exclusively on labeled data, i.e., \textit{when} do graph pre-training models work on GAD? 
(2) And the insights under which pre-training is most effective, i.e., \textit{why} do graph pre-training models work on GAD?

In this work, we take the initiative to systematically analyze the \textit{when and why} of employing pre-training models in the context of GAD.
Through extensive experiments against a broad spectrum of leading baselines, we uncover that, pre-training models, even employed with vanilla GNN backbone, consistently exhibit extraordinarily competitive performances across various real-world datasets.
Remarkably, pre-training models significantly surpass current state-of-the-art models, including those with intricate architectures or advanced classifiers, by a significant margin under scenarios of limited supervision.

Our analysis further explores the mechanisms that underpin the effectiveness of pre-training models in GAD, uncovering a strong correlation with graph density.
Motivated by these findings, we examine the dynamics of label information propagation and empirically confirm that the enhanced ability of pre-training models to detect under-represented anomalies, located beyond 2-hop neighborhoods of known anomalies, is a key factor distinguishing them from end-to-end trained models.
Furthermore, we provide empirical evidence that the negative sampling can be regarded as the generation of `pseudo anomalies', which, through contrastive learning, aligns with the GAD objective and improves detection capabilities in downstream tasks.
However, these relative benefits from the pre-training stage will diminish with increasing supervision ratio.
At last, we explore the potential of pre-training in graph-level anomaly detection, a notably more challenging anomaly detection task within the realm of graph-structured data.

In a nutshell, the main contributions of this work are as follows:
\begin{itemize}
    \item To the best of our knowledge, we are the first work to systematically analyze the effectiveness of pre-training in the context of GAD, exploring \textit{when} and \textit{why} graph pre-training models work for anomaly detection.
    \item We provide empirical evidence highlighting the significant advantages of employing pre-training models in GAD, particularly under conditions characterized by graphs of lower density and limited supervision.
    \item Through rigorous empirical analyses of diverse real-world datasets, we uncover the factors that impact the success of pre-training in GAD, such as the detection of under-represented anomalies, mechanism of pre-training pretext, and level of supervision.
    These analyses provide valuable insights into the future integration of pre-training in GAD tasks.
\end{itemize}

\section{Related Works}

\subsection{Graph Anomaly Detection}

Leveraging deep learning techniques in GAD has become a focal point of recent research, yielding substantial progress. 
Ma et al. \cite{anomaly_survey} comprehensively reviewed the use of deep learning for detecting anomalies like nodes, edges, and subgraphs in both static and dynamic graphs. Benchmarking tools such as BOND \cite{BOND} in unsupervised settings and GADBench \cite{tang2023gadbench} in supervised environments have been crucial in evaluating various GAD methodologies. 
Additionally, a multitude of surveys have investigated GAD's practical applications in diverse contexts, including the identification of anomalous accounts and bots in social networks \cite{anand2017anomaly}, sensor faults in IoT networks \cite{gaddam2020detecting}, fake news in social media \cite{aimeur2023fake, bondielli2019survey}, and financial fraud in transaction networks \cite{hilal2022financial}.

GNNs have recently gained popularity for mining graph data~\cite{GAT,GCN,GIN,GraphSAGE}. 
To address the specific challenges of graph anomalies, numerous adaptations of standard GNNs have been proposed \cite{GAD_timeseries,zheng2019addgraph,GRADATE,MAG,mulgad,ding2019deep,wang2019semi,geniepath,zhang2021fraudre}. 
Innovations such as GraphConsis \cite{liu2020alleviating} and CARE-GNN \cite{CareGNN} have been developed to counteract the camouflage behavior of anomalies through enhanced message passing and aggregation processes. 
Approaches like PC-GNN \cite{PCGNN} and DAGAD \cite{DAGAD} tackle the issue of label imbalance by employing techniques like imbalance-aware data sampling and graph augmentation, emphasizing the weight of anomalies during training. 
The introduction of spectral GNNs \cite{BWGNN, GHRN} has marked a significant advancement, associating anomalies with high-frequency spectral patterns and employing versatile frequency filters to better capture the anomalies' signals. 
Despite these advancements, current supervised methods predominantly focus on developing effective GNN encoders for node representation learning based on labels, with other potential pre-training models remaining largely unexplored.

\begin{table*}[ht]
    \centering 
    \caption{Statistics of graph anomaly detection datasets.}
    \label{table:node_datasets}
    \begin{tabular}{cccccccc}
    \toprule
    Dataset & \# Nodes & \# Edges & Anomaly & Density & Avg. Deg. & Avg. Deg. (Anomaly) & Relation Concept \\
    \midrule
    Reddit & 10,984 & 168,016 & 3.3\% & 0.14\% & 15.3 & 12.4 & Under Same Post \\
    Weibo & 8,405 & 407,963 & 10.3\% & 0.55\% & 48.5 & 27.6 & Under Same Hashtag \\
    Amazon & 11,944 & 4,398,392 & 9.5\% & 3.08\% & 368.3 & 268.1 & Review Correlation \\
    YelpChi & 45,954 & 3,846,979 & 14.5\% & 0.18\% & 83.7 & 160.2 & Reviewer Interaction \\
    Tolokers & 11,758 & 519,000 & 21.8\% & 0.38\% & 44.1 & 136.3 & Work Collaboration \\
    Questions & 48,921 & 153,540 & 3.0\% & <0.01\% & 3.1 & 18.9 & Question Answering \\
    T-Finance & 39,357 & 21,222,543 & 4.6\% & 1.37\% & 539.2 & 652.5 & Transaction Record \\
    Elliptic & 203,769 & 234,355 & 9.8\% & <0.01\% & 1.2 & 1.3 & Payment Flow \\
    DGraph-Fin & 3,700,550 & 4,300,999 & 1.3\% & <0.01\% & 1.2 & 1.2 & Loan Guarantor \\
    T-Social & 5,781,065 & 73,105,508 & 3.0\% & <0.01\% & 12.6 & 173.2 & Social Friendship \\
    \bottomrule
    \end{tabular}
\end{table*}

\subsection{Graph Pre-Training}

Graph pre-training models first learn universal knowledge from unlabeled data with self-supervised objectives, and then transfer the knowledge to deal with specific downstream tasks.
According to the design of self-supervised objectives, graph pre-training can be categorized into two main groups: contrastive learning and predictive learning~\cite{liu2021graph, xie2022self}. 
Contrastive learning aims to align instances from multi-views through mutual information maximization.
The majority of the research has been dedicated to the design of negative sampling and augmentation schemes, such as corruptions in DGI~\cite{velivckovic2018deep}, graph diffusion in MVGRL~\cite{hassani2020contrastive}, masking in GRACE~\cite{zhu2020grace}, latent noise in COSTA~\cite{zhang2022costa}, and spectral augmentation in SFA~\cite{zhang2023spectral}.
Some studies have attempted negative-sample-free learning methods. 
For example, BGRL~\cite{thakoor2022bgrl} employs strong regularization from architecture designs, and CCA-SSG~\cite{zhang2021ccassg} conducts feature decorrelation.
In contrast, predictive learning learns informative properties by information reconstruction for pretext tasks, typically using autoencoder~\cite{hinton1993autoencoders}.
Early works like VGAE~\cite{kipf2016vgae} and ARVGA~\cite{pan2018arvga} utilize structural reconstruction.
More recent research has shifted focus towards the design of decoders, including re-masking in GraphMAE\cite{hou2022graphmae, hou2023graphmae2}, spectral decoder in WGDN~\cite{cheng2023wiener}, and cross-correlation decoding in S2GAE~\cite{tan2023s2gae}.

Existing literature on graph pre-training has largely centered around its application in general classification tasks. 
A recent study by DCI~\cite{wang2021decoupling} embarks on an initial investigation into applications of pre-training in GAD, arguing its enhanced ability to identify `hard instances' that end-to-end models often struggle to identify
Nonetheless, there still exists a discernible lack of comprehensive and systematic examination of pre-training's role and efficacy in the GAD landscape.

\section{Preliminaries}

In this section, we begin by formalizing the problem of detecting anomalies in graphs and then proceed to define two paradigms for solving this problem: the end-to-end training model and the pre-training model.

In a general scenario, we are given a static attributed graph $\mathcal{G} = (\mathcal{V}, \mathbf{A}, \mathbf{X})$, which comprises the following components: 
(1) $\mathcal{V} = \{ v_{1}, v_{2}, ..., v_{N} \}$ is the set of nodes; 
(2) $\mathbf{A} \in \mathbb{R}^{N \times N}$ is the adjacency matrix where $\mathbf{A}_{ij} \in \{0, 1\}$ indicates the presence or absence of an edge between $v_i$ and $v_j$; and 
(3) $\mathbf{X} \in \mathbb{R}^{N \times D}$ denotes the feature matrix, where $x_{i}$ denotes the $d$-dimensional feature vector of node $v_{i}$. 

\vpara{Graph anomaly detection.}
Let $\mathcal{V}_{a}$ and $\mathcal{V}_{n}$ be two disjoint subsets of $\mathcal{V}$, where $\mathcal{V}_{a}$ consists of labeled anomalous nodes and $\mathcal{V}_{n}$ comprises labeled normal nodes.
Given a partially labeled attributed graph $\mathcal{G} = (\mathcal{V}, \mathbf{A}, \mathbf{X})$ with the set of partial labels $\mathbf{Y}^{L}$ of $\mathcal{V}^{L} = \mathcal{V}_{a} \cup \mathcal{V}_{n}$, our objective is to learn a predictive function
\begin{equation}
    \mathcal{F} : \mathcal{G} = (\mathcal{V}, \mathbf{A}, \mathbf{X}) \mapsto \mathbf{Y}
\end{equation}
to identify the anomalous status $\mathbf{Y}^{U} = \mathbf{Y} \setminus \mathbf{Y}^{L}$ of unlabeled nodes in $\mathcal{V}^{U} = \mathcal{V} \setminus \mathcal{V}^{L}$.
Note that authentic labels are often expensive to obtain, we assume that the label information is only available for a small number of nodes (i.e. $|\mathcal{V}^{L}| \ll |\mathcal{V}|$).
Furthermore, there are usually significantly fewer anomalous nodes than normal nodes (i.e. $|\mathcal{V}_{a}| \ll |\mathcal{V}_{n}|$).
Given such, GAD can be viewed as an imbalanced binary node classification problem, but its main focus is to identify unusual and deviated behaviors.

In convention, the predictive function $\mathcal{F}$ is usually divided in to a GNN encoder $g_{\theta}: \mathcal{G} \mapsto \mathbb{R}^{d'}$ and a binary classifier $f_{\phi}: \mathbb{R}^{d'} \mapsto \{ 0, 1\}$, where $\theta$ and $\phi$ are the parameters to be optimized.
Specifically, $g_{\theta}$ is to encode the structure patterns into the node representations $\mathbf{H} = g_{\theta}(\mathcal{G})$, and $f_{\phi}$ is applied on top of the node representations to distinguish anomalous status $\hat{\mathbf{Y}} = f_{\phi}(\mathbf{H})$.
Based on the parameter optimization schemes, the learning paradigms of current GNN-based anomaly detection models can be classified into two categories.

\vpara{End-to-End training model.} 
In general, an end-to-end training model jointly optimizes the encoder $g_{\theta}$ and classifier $f_{\phi}$ by minimizing the discrepancy between predictions $\hat{\mathbf{Y}} = f_{\phi} \, \circ \, g_{\theta}(\mathcal{G})$ and ground truth $\mathbf{Y}$, using a supervised loss function $\mathcal{L}_{SL}$.

\vpara{Pre-training model.}
For a pre-training model, the optimization process of encoder $g_{\theta}$ and classifier $f_{\phi}$ is decoupled into two stages: pre-training and fine-tuning.
In the pre-training stage, the encoder $g_{\theta}$ is optimized using a self-supervised loss function $\mathcal{L}_{SSL}$ for pretexts tasks that do not rely on supervision information $\mathbf{Y}^{L}$.
Subsequently, the classifier $f_{\phi}$ is fine-tuned using a supervised loss function $\mathcal{L}_{SL}$ while keeping the learned encoder $g_{\theta}$ frozen.

\begin{table*}[t]
    \centering 
    \caption{Comparison of the AUROC score of each model in semi-supervised settings. 
    The best and runner-up models are highlighted in bold and underlined. Results are averaged across 10 runs. 
    The \textcolor{orange}{orange}/\textcolor{blue}{blue} color indicates that pre-training outperforms/underperforms the counterpart.}
    \label{table:node_semi_auroc}
    \resizebox{\textwidth}{!}{
    \begin{tabular}{c|cccccccccc|c}
    \toprule
    Model & Reddit & Weibo & Amazon & YelpChi & T-Finance & Elliptic & Tolokers & Questions & DGraph-Fin & T-Social & Average \\
    \midrule
    GCN & 60.37 & 96.13 & 85.14 & 54.66 & 90.90 & 86.96 & 68.32 & 65.03 & 65.32 & 85.22 & 75.90 \\
    GIN & 60.03 & 83.82 & 91.61 & 64.46 & 84.51 & 88.24 & 66.84 & 62.21 & 65.73 & 70.42 & 73.79 \\
    PCGNN & 52.81 & 83.92 & 93.22 & 65.12 & 92.03 & 87.55 & 67.43 & 59.02 & 68.43 & 69.12 & 73.87 \\
    GAT-sep & 60.31 & 87.82 & 91.42 & 65.02 & 86.31 & 89.31 & 69.12 & 61.92 & \textbf{69.03} & 74.51 & 75.48 \\
    BWGNN & 57.72 & 93.63 & 93.15 & 65.31 & 92.12 & 88.73 & 68.51 & 63.60 & 68.40 & 77.51 & 76.87 \\
    GHRN & 57.51& 91.62 & 90.92 & 64.51 & 92.63 & 89.02 & 69.01 & 60.52 & 67.12 & 78.73 & 76.16 \\
    RF-Graph & \underline{61.43} & 96.32 & 92.51 & 61.63 & \textbf{95.04} & \textbf{93.81} & 70.42 & 65.71 & 64.40 & 88.63 & 78.89 \\
    XGB-Graph & 59.23 & 97.42 & \textbf{94.71} & 64.08 & \underline{94.82} & \underline{90.92} & 67.54 & 61.41& 62.36 & 85.20 & 77.77 \\
    \midrule
    GraphMAE & 61.36 & \textbf{98.71} & 92.62 & \underline{68.07} & 92.63 & 88.41 & \textbf{71.63} & \underline{68.12} & 65.74 & \underline{90.54} & \underline{79.78} \\
    DGI & \textbf{64.53} & \underline{98.53} & \underline{93.51} & \textbf{68.92} & 91.41 & 89.49 & \underline{71.49} & \textbf{70.71} & \underline{68.43} & \textbf{91.38} & \textbf{80.84} \\
    Imp. Back. & \textcolor{orange}{+ 4.16} & \textcolor{orange}{+ 2.57} & \textcolor{orange}{+ 1.90} & \textcolor{orange}{+ 4.46} & \textcolor{orange}{+ 1.73} & \textcolor{orange}{+ 1.25} & \textcolor{orange}{+ 3.31} & \textcolor{orange}{+ 5.68} & \textcolor{orange}{+ 2.70} & \textcolor{orange}{+ 6.16} & \textcolor{orange}{+ 4.94} \\ 
    Imp. SOTA & \textcolor{orange}{+ 3.10} & \textcolor{orange}{+ 1.29} & \textcolor{blue}{- 1.20} & \textcolor{orange}{+ 3.61} & \textcolor{blue}{- 2.41} & \textcolor{blue}{- 4.32} & \textcolor{orange}{+ 1.21} & \textcolor{orange}{+ 5.00} & \textcolor{blue}{- 0.60} & \textcolor{orange}{+ 2.75} & \textcolor{orange}{+ 1.95} \\
    \bottomrule
    \end{tabular}
    }
\end{table*}
\begin{table*}[t]
    \centering 
    \caption{Comparison of the AUROC score of each model in fully-supervised settings. 
    The best and runner-up models are highlighted in bold and underlined. Results are averaged across 10 runs.
    The \textcolor{orange}{orange}/\textcolor{blue}{blue} color indicates that pre-training outperforms/underperforms the counterpart.}
    \label{table:node_full_auroc}
    \resizebox{\textwidth}{!}{
    \begin{tabular}{c|cccccccccc|c}
    \toprule
    Model & Reddit & Weibo & Amazon & YelpChi & T-Finance & Elliptic & Tolokers & Questions & DGraph-Fin & T-Social & Average \\
    \midrule
    GCN & 62.04 & 98.84 & 85.17 & 58.62 & 94.62 & 81.69 & 73.80 & 68.20 & 75.51 & 96.63 & 79.51 \\
    GIN & 61.82 & 98.65 & 95.63 & 73.77 & 92.71 & 83.01 & 74.57 & 68.08 & 74.16 & 94.09 & 81.65 \\
    PCGNN & 65.41 & 95.14 & 98.01 & 80.82 & 94.03 & 86.50 & 76.63 & 67.66 & 72.76 & 96.91 & 83.38  \\
    GAT-sep & 66.17 & 98.22 & 95.11 & 80.49 & 94.40 & 85.94 & 79.52 & 70.08 & 75.78 & 87.69 & 83.33 \\
    BWGNN & \underline{70.82} & 98.13 & 98.27 & 87.13 & 96.93 & 87.03 & 80.41 & 70.87 & 76.30 & 96.88 & 86.28 \\
    GHRN & 61.02 & 99.18 & \underline{98.29} & 84.60 & 96.46 & 89.50 & 80.08 & 72.16 & 76.13 & 97.12 & 85.46 \\
    RF-Graph & 65.35 & \textbf{99.43} & 96.73 & \underline{95.24} & \textbf{97.28} & \textbf{93.21} & \underline{81.88} & 64.77 & 67.78 & \underline{99.69} & 86.14 \\
    XGB-Graph & 64.74 & \underline{99.29} & \textbf{98.74} & \textbf{97.37} & \underline{97.15} & \underline{91.78} & \textbf{82.85} & 71.02 & 75.83 & \textbf{99.76} & \textbf{87.85} \\
    \midrule
    GraphMAE & 68.19 & 99.21 & 97.69 & 86.95 & 94.92 & 85.24 & 79.85 & \underline{74.50} & \underline{76.32} & 96.55 & 85.94 \\
    DGI & \textbf{71.84} & 99.24 & 97.61 & 86.58 & 94.71 & 87.97 & 80.24 & \textbf{74.64} & \textbf{76.81} & 97.25 & \underline{86.69} \\
    Imp. Back. & \textcolor{orange}{+ 9.80} & \textcolor{orange}{+ 0.40} & \textcolor{orange}{+ 2.06} & \textcolor{orange}{+ 13.18} & \textcolor{orange}{+ 0.30} & \textcolor{orange}{+ 4.96} & \textcolor{orange}{+ 5.67} & \textcolor{orange}{+ 6.44} & \textcolor{orange}{+ 1.30} & \textcolor{orange}{+ 0.62} & \textcolor{orange}{+ 5.04} \\
    Imp. SOTA & \textcolor{orange}{+ 1.02} & \textcolor{blue}{- 0.19} & \textcolor{blue}{- 1.04} & \textcolor{blue}{- 10.42} & \textcolor{blue}{- 2.36} & \textcolor{blue}{- 5.23} & \textcolor{blue}{- 2.61} & \textcolor{orange}{+ 2.48} & \textcolor{orange}{+ 0.51} & \textcolor{blue}{- 2.51} & \textcolor{blue}{- 1.16} \\
    \bottomrule
    \end{tabular}
    }
\end{table*}

\section{Experimental Results}

The following section details our thorough evaluation.
We first introduce evaluation configurations, followed by a systematic examination of scenarios where pre-training optimally improves GAD.

\subsection{Configurations}

\vpara{Datasets}
Following the most recent benchmark GADBench~\cite{tang2023gadbench}, here we integrated a large number of datasets across various scales and domains.
Detailed statistics are presented in Table~\ref{table:node_datasets}.
Among these datasets, Weibo~\cite{zhao2020error,BOND}, Reddit~\cite{kumar2019predicting,BOND}, Questions~\cite{HeteroBench}, and T-Social~\cite{BWGNN} are designed to detect anomalous accounts in social media networks. 
Tolokers~\cite{HeteroBench}, Amazon~\cite{mcauley2013amateurs,CareGNN}, and YelpChi~\cite{rayana2015collective,CareGNN} aim to identify fraudulent workers, reviews, and reviewers on crowd-sourcing or e-commerce platforms.
T-Finance~\cite{BWGNN}, Elliptic~\cite{weber2019anti}, and DGraph-Fin~\cite{dgraph_dataset} are financial networks containing fraudulent users, illicit entities, and overdue loans.

\vpara{Baselines.}
The baselines can be categorized into two groups.
The first group consists of end-to-end learning models.
This group includes standard GNNs such as GCN~\cite{GCN}, GIN~\cite{GIN}, GNNs specifically tailored for GAD including PC-GNN~\cite{PCGNN}, GAT-sep~\cite{GATsep}, BWGNN~\cite{BWGNN}, GHRN~\cite{GHRN}, RF-Graph, XGB-Graph~\cite{tang2023gadbench}.
The second group comprises pre-training models.
We select the representative models for both predictive and contrastive learning, including GraphMAE~\cite{hou2022graphmae} and DGI~\cite{velivckovic2018deep}.

\vpara{Evaluation protocols.}
In alignment with the existing anomaly detection benchmarks~\cite{zhao2021using,han2022adbench,tang2023gadbench}, we evaluate different methods leveraging two widely used metrics: Area Under Receiver Operating Characteristic Curve (AUROC) and Area Under Precision-Recall Curve (AUPRC).
Among these metrics, AUROC gauges the overall performance while AUPRC is more responsive to the prediction ranking of anomalous nodes.
To ensure a thorough evaluation, we explore both semi-supervised and fully-supervised scenarios.
For semi-supervised settings, the training set across all datasets contains 20 anomalous nodes and 80 normal nodes.
For fully-supervised settings, we adopt the configurations established by GADBench. 
To ensure robustness, we conduct 10 trials with different splits and report the average performance.
To avoid the potential impact of advanced classifiers, we select trivial linear classifiers for downstream evaluation.
Specifically, the embeddings learned in the pre-training stage are frozen and then fed into a 2-layer MLP~\cite{MLP} for predictions.

\vpara{Implementation details.}
For all baselines, we use the official implementation provided by the authors.
To mitigate the impact of GNNs' architecture, we exclusively employ standard GNNs, specifically GCN and GIN, as the backbone candidates for pre-training.
For each dataset, we choose the better-performing model between GCN and GIN to serve as the backbone architecture.
To ensure fairness in hyperparameter tuning, we utilize the grid search within the predefined search space to optimize hyperparameters.
Additional information regarding model configurations, hyperparameter optimization, and other implementation specifics can be found in Appendix~\ref{sec:appendix_exp_setup}.

\subsection{When Pre-Training Excels in Graph Anomaly Detection}

In Table~\ref{table:node_semi_auroc} and~\ref{table:node_full_auroc}, we take a close look at the model performance regarding the AUROC score after hyper-parameter tuning on each dataset.
For additional experimental results, please refer to Appendix~\ref{sec:appendix_exp_setup}.
Our primary findings include:

\vpara{Pre-training has superior performance with limited supervision.}
As illustrated in Table~\ref{table:node_semi_auroc} and~\ref{table:node_semi_auprc}, pre-training with basic GNN backbones yields notable achievements.
In semi-supervised settings, the two variants of pre-training methods, namely GraphMAE and DGI, stand out by achieving state-of-the-art performances in comparison to other end-to-end learning baselines across 6 out of 10 datasets.
In particular, pre-training consistently surpasses its backbone model trained in end-to-end manner over all datasets, showcasing an absolute average improvement of 4.94\% on AUROC, and 8.83\% on AUPRC.
The emphasis on AUPRC improvement against backbones is particularly noteworthy, suggesting that pre-training enhances the capability to predict high-confidence anomalies.
In addition, DGI presents an absolute average improvement of 1.95\% on AUROC and 0.25\% on AUPRC when compared to RF-Graph, the best end-to-end learning model in this setting, which employs the advanced random forest classifier.
In comparison to top-performing baselines that utilize linear classifiers, BWGNN, DGI demonstrate a significantly more substantial enhancement, with an absolute growth of 3.97\% on AUROC and 8.11\% on AUPRC.
Even when evaluated against the best end-to-end models for each dataset, pre-training exhibits variability in performance, yet it maintains an overall absolute average improvement of 0.84\% on AUROC.

\vpara{Pre-training achieves comparable performance to the SOTA model with sufficient supervision.}
As depicted in Table~\ref{table:node_full_auroc} and~\ref{table:node_full_auprc}, pre-training methods consistently maintain competitiveness when compared to end-to-end learning methods in fully-supervised settings. 
Significantly, they demonstrate an average AUROC performance on par with XGB-Graph, the leading end-to-end learning model in this setting.
Particularly, DGI continues to achieve state-of-the-art performances in 3 datasets.
Simultaneously, DGI outperforms BWGNN, the leading baseline model utilizing linear classifiers, with an average absolute increase of 0.41\% on AUROC and 1.77\% on AUPRC.
Furthermore, it is essential to highlight that even in situations of sufficient supervision, pre-training outperforms its backbone model across all datasets, demonstrating an absolute average improvement of 6.13\% on AUROC, and 12.14\% on AUPRC.
Importantly, the improvements are more pronounced than those observed in semi-supervised settings, indicating the greater benefits conferred upon a basic GNN backbone through pre-training when provided with more supervision.

\vpara{Summary.}
Drawing upon the results discussed, we summarize that the integration of pre-training, even when applied to a basic GNN backbone, consistently yields improvements in the task of GAD.
These findings undeniably affirm the effectiveness of pre-training in the context of GAD.
The pronounced benefits observed in semi-supervised settings further suggest that pre-training models optimally enhance GAD under the conditions of limited supervision.

\begin{table}[t]
    \centering
    \caption{Average improvements on AUROC score towards backbone and SOTA models on graphs with different densities.
    \textcolor{orange}{\large$\blacktriangle$}/\textcolor{blue}{\huge$\blacktriangledown$} indicates that pre-training outperforms/underperforms the counterpart.}
    \label{table:data_category}
    \begin{tabular}{c|c|cc}
    \toprule
    Setting & Density & Avg. Imp. Back. & Avg. Imp. SOTA \\
    \midrule
    \multirow{3}{*}{Semi} & Sparse & 4.39\textcolor{orange}{\large$\blacktriangle$} & 3.62\textcolor{orange}{\large$\blacktriangle$} \\
    & Dense & 1.82\textcolor{orange}{\large$\blacktriangle$} & 0.71\textcolor{blue}{\huge$\blacktriangledown$} \\
    & Over-Sparse & 1.98\textcolor{orange}{\large$\blacktriangle$} & 0.15\textcolor{blue}{\huge$\blacktriangledown$} \\
    \midrule
    \multirow{3}{*}{Fully} & Sparse & 5.97\textcolor{orange}{\large$\blacktriangle$} & 0.81\textcolor{blue}{\huge$\blacktriangledown$} \\
    & Dense & 1.09\textcolor{orange}{\large$\blacktriangle$} & 1.79\textcolor{blue}{\huge$\blacktriangledown$} \\
    & Over-Sparse & 1.88\textcolor{orange}{\large$\blacktriangle$} & 1.42\textcolor{blue}{\huge$\blacktriangledown$} \\
    \bottomrule
    \end{tabular}
\end{table}

\section{Why Pre-Training Excels in Graph Anomaly Detection}

To gain deeper insights, in this section, we switch our attention to addressing a vital question: \textit{why pre-training excels in graph anomaly detection?}

\subsection{The Impact of Graph Sparsity} 

While pre-training models generally demonstrate promising performance, they exhibit variability in performance across different datasets.
To gain insights into this variability, we first analyze the characteristics of the datasets 
As depicted in Table~\ref{table:node_datasets}, the most substantial differences among these datasets lie in their graph density and average node degrees.
Based on these properties, we categorize the datasets into three types: \textit{Sparse Graphs}, \textit{Dense Graphs}, and \textit{Over-Sparse Graphs}.
\textit{Sparse Graphs} denote datasets with graph density below 1\% and average node degrees above 2, which includes Reddit, Weibo, YelpChi, Tolokers, Questions, and T-Social.
\textit{Dense Graphs} refer to the dataset with graph density exceeding 1\%, such as Amazon and T-Finance.
\textit{Over-Sparse Graphs} consist of Elliptic and DGraph-Fin characterized by extremely low graph density, resulting in average node degrees of around 1.

To better investigate the variations in pre-training methods'' performance across datasets, we present the average improvements in AUROC within each category relative to both the backbone and best baseline models, namely RF-Graph in semi-supervised and XGB-Graph in fully-supervised settings, as shown in Table~\ref{table:data_category}.
These results highlight that pre-training methods consistently deliver significant improvements over the baseline model across all scenarios.
Notably, the most significant improvements are observed in sparse graphs, particularly in fully-supervised settings.
When compared to the SOTA model, we observe that when limited supervision is available, pre-training methods achieve significant positive improvements on sparse graphs but fall short on the remaining datasets.
Furthermore, even in fully-supervised settings, pre-training methods present a marginal decrease of 0.81\% on AUROC for sparse graphs, which is an improvement compared to the overall absolute decrease of 1.16\% as presented in Table~\ref{table:node_full_auroc}.
These empirical findings strongly suggest that pre-training methods tend to excel on graphs with lower density, prompting further investigation into the underlying factors driving these enhancements.

\begin{figure}[t]
    \centering
    \includegraphics[width=\linewidth]{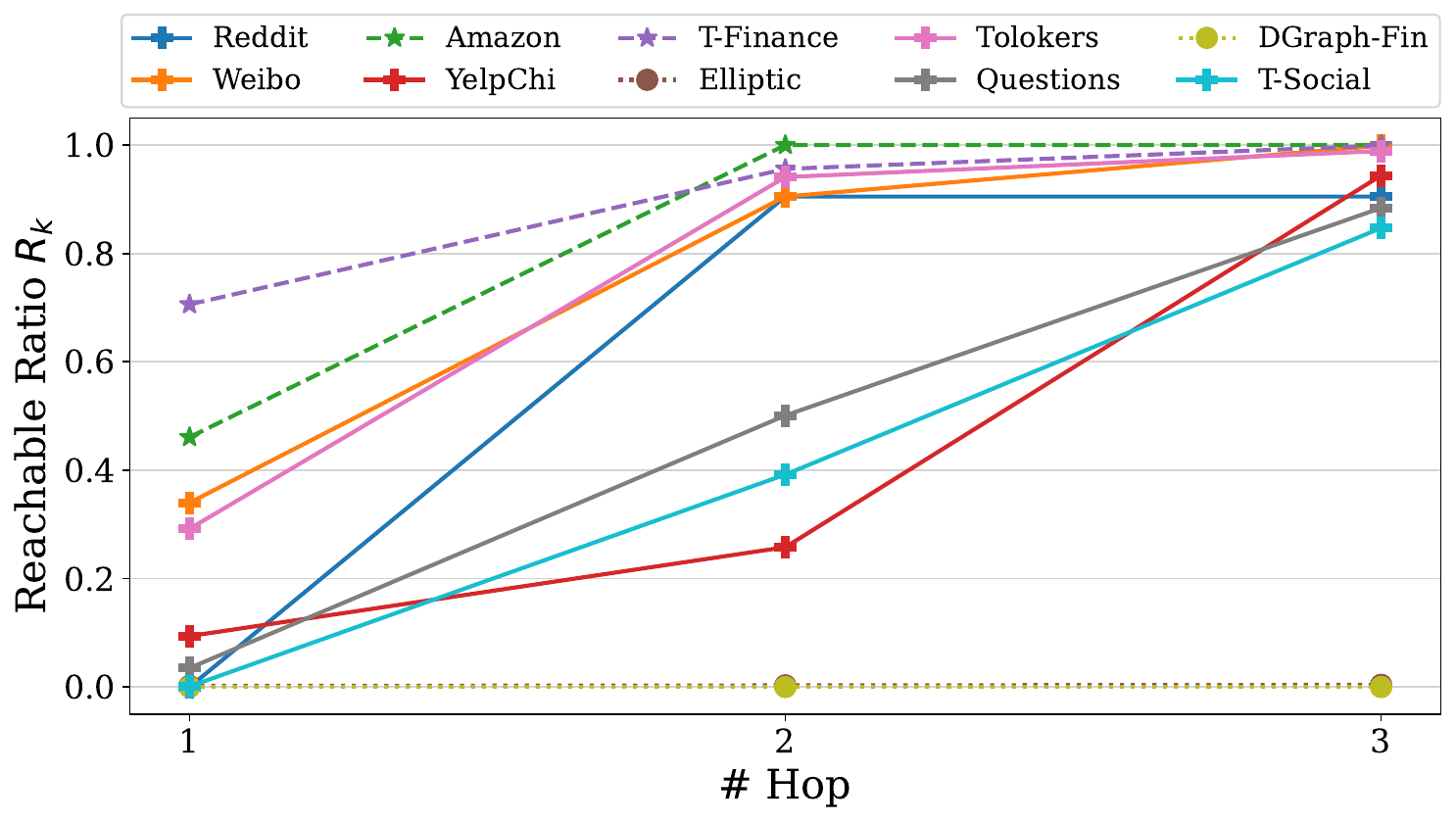}
    \caption{The k-hop reachable ratio $R_{k}$ till 3-hop neighborhood across all datasets.}
    \label{fig:hop_reach}
\end{figure}

\subsection{From Sparsity to Reachable Ratio}

Building upon previous investigations, our focus is primarily on semi-supervised settings where pre-training offers the most substantial advantages.
We first explore how low graph density influences the usefulness of pre-training methods.

A previous study SLAPS~\cite{fatemi2021slaps} argues that lower graph density results in more edges receiving no supervision from labeled nodes, thus presenting increased challenges when training GNNs in an end-to-end manner.
Motivated by these insights, we propose to investigate this problem from a more fundamental perspective, specifically focusing on the propagation of label information.
Notably, in the context of GAD, normal instances significantly outnumber anomalies.
However, our primary goal is to identify anomalous nodes, necessitating greater attention to the label information of these anomalies. 
As a result, we introduce a novel metric to assess the feasibility of propagating information from anomalous nodes to unlabelled ones. 
This metric is formally defined as follows:
\begin{definition}[K-hop reachable ratio]
    Given the set of labeled anomalous nodes $\mathcal{V}_{a}$, the k-hop reachable ratio is defined as 
    \begin{equation}
        R_{k} = \frac{|u \in (\mathcal{N}_{k}(\mathcal{V}_{a}) \cap \mathcal{V}^{U}_{a})|}{|\mathcal{V}^{U}_{a}|},
    \end{equation}
    where $\mathcal{N}_{k}(\mathcal{V}_{a})$ denotes the union of the node set within the k-hop neighborhood of each $v_{i} \in \mathcal{V}_{a}$ and $\mathcal{V}_{a}^{U}$ represents the set of all unlabeled anomalies. 
\end{definition}

In essence, the k-hop reachable ratio $R_{k}$ quantifies the proportion of unlabeled anomalies reachable within the k-hop neighborhood of labeled ones
A higher $R_{k}$ implies a greater likelihood of propagating information from labeled anomalies to unlabeled ones.
It's important to note that $R_{k}$ primarily depends on graph density and the quantity of labeled anomalies.
Specifically, given a fixed number of labeled anomalies, a lower graph density presents greater challenges in accessing more anomalous nodes within the neighborhoods, consequently resulting in a lower $R_{k}$.

\subsection{Pre-training Improves the Detection of 2-hop Unreachable Anomalies}

To explore the association between k-hop reachable ratios and the effectiveness of pre-training methods, we initiate our analysis by assessing $R_{k}$ across various datasets within our semi-supervised framework, as illustrated in Figure~\ref{fig:hop_reach}.
Notably, for Elliptic and DGraph-Fin, we observe no significant deviations in the reachable ratios.
We attribute this phenomenon to the extreme sparsity of these datasets.
For the remaining datasets, it can be observed that the deviations in reachable ratios primarily occur within 1-hop and 2-hop neighborhoods.
For sparse graphs, at least one of $R_{1}$ and $R_{2}$ displays significantly lower values than those of denser ones such as T-Finance.
In addition, integrating the results presented in Table~\ref{table:node_semi_auroc}, we find that pre-training methods demonstrate particularly pronounced improvements on datasets with $R_{2}$ lower than 0.5, such as YelpChi, Questions, and T-Social, which reports absolute increase in AUROC of 3.61\%, 5.00\% and 2.75\% against SOTA models, respectively.
These findings suggest that the 2-hop reachable ratio $R_{2}$ may serve as a pivotal factor in contributing to the superior performance of pre-training methods.

\begin{figure}
    \subfigure{
    \centering
    \includegraphics[width=0.47\columnwidth]{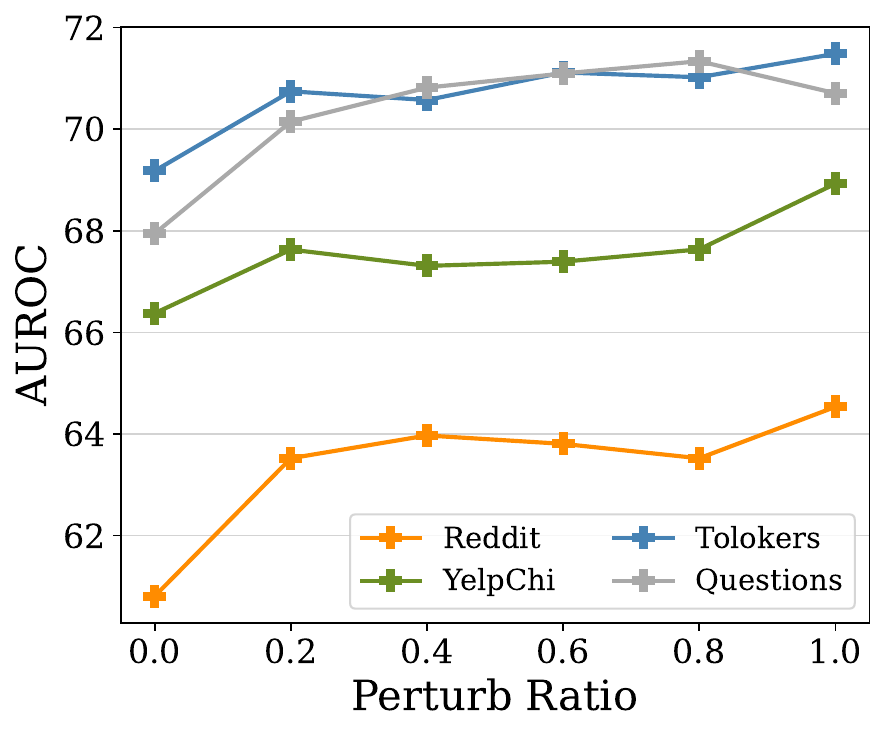}
    }
    \subfigure{
    \centering
    \includegraphics[width=0.47\columnwidth]{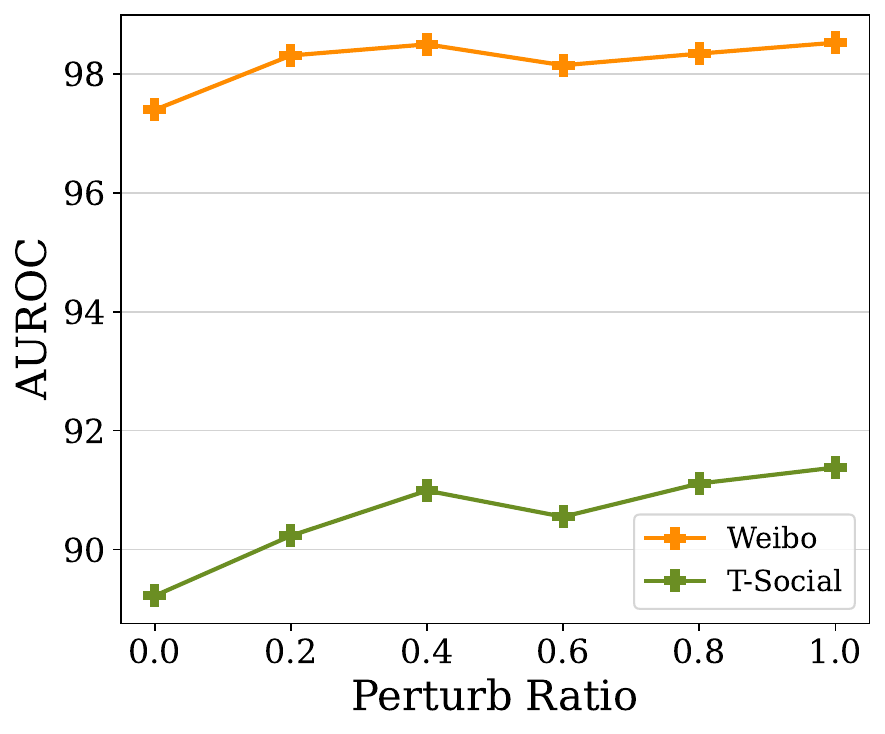}
    }
    \caption{The average AUROC score of DGI in downstream anomaly detection with different perturbation ratios used in the pre-training stage.
    \label{fig:perturb_ratio}}
\end{figure}

As previously discussed, a lower $R_{2}$ signifies not only a reduced access to unlabeled anomalies but also a constrained propagation of information from labeled anomalies.
Consequently, there's a higher risk of insufficient label signals being transmitted to unlabeled anomalies beyond 2-hop neighborhoods when GNNs are trained in an end-to-end manner.
Building on this understanding, we posit that the superior performance of pre-training models primarily stems from their ability to detect 2-hop unreachable anomalies in scenarios with limited supervision.
To validate our hypothesis, we explore how different models perform in identifying anomalous nodes across various hop neighborhoods.

To avoid ambiguity, we first categorize the anomalies in test time based on their shortest path to the nearest labeled anomalies.
For instance, a 2$^{nd}$-hop anomaly is one located in the 2$^{nd}$-hop of the nearest labeled anomaly.
To comprehensively gauge the model capability, we conduct evaluations on four representative sparse graphs with varying $R_{2}$ values, including YelpChi, Questions, T-Social, and Tolokers.
We compare DGI with its backbone as well as the SOTA end-to-end learning baselines for each dataset.
It's important to note that the optimal classification threshold differs for each model. 
To mitigate this potential variability, we evaluate the anomaly detection ability by the average prediction ranking of k$^{th}$-hop anomalies.

\begin{figure*}[!h]
    \subfigure[YelpChi ($R_{2}$=0.26)]{
    \centering
    \begin{minipage}[b]{0.23\textwidth}
    \includegraphics[width=1.0\textwidth]{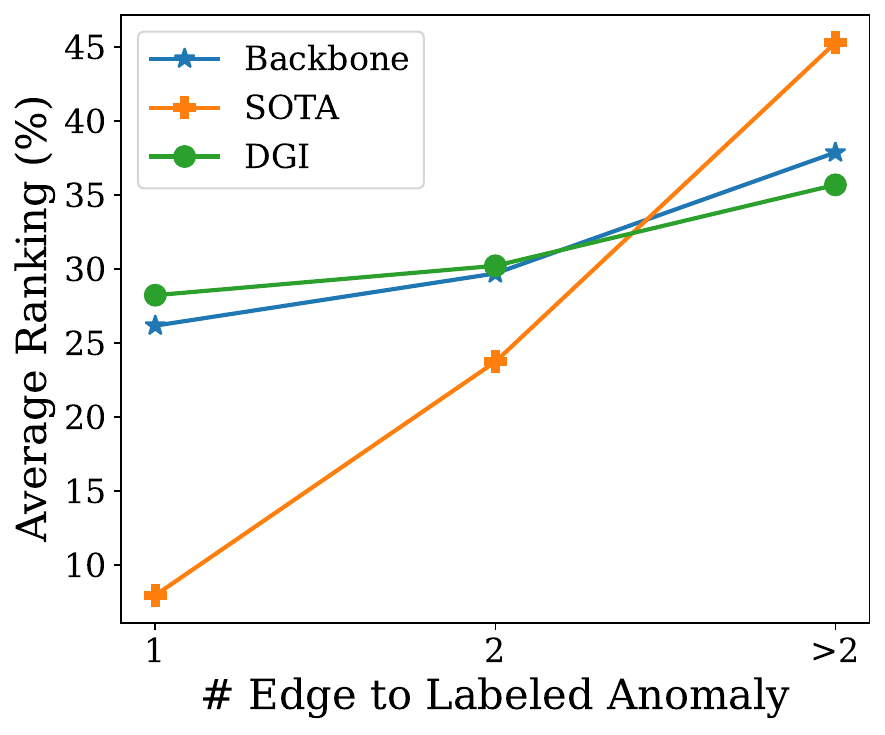}
    \end{minipage}
    }
    \subfigure[Questions ($R_{2}$=0.50)]{
    \centering
    \begin{minipage}[b]{0.23\textwidth}
    \includegraphics[width=1.0\textwidth]{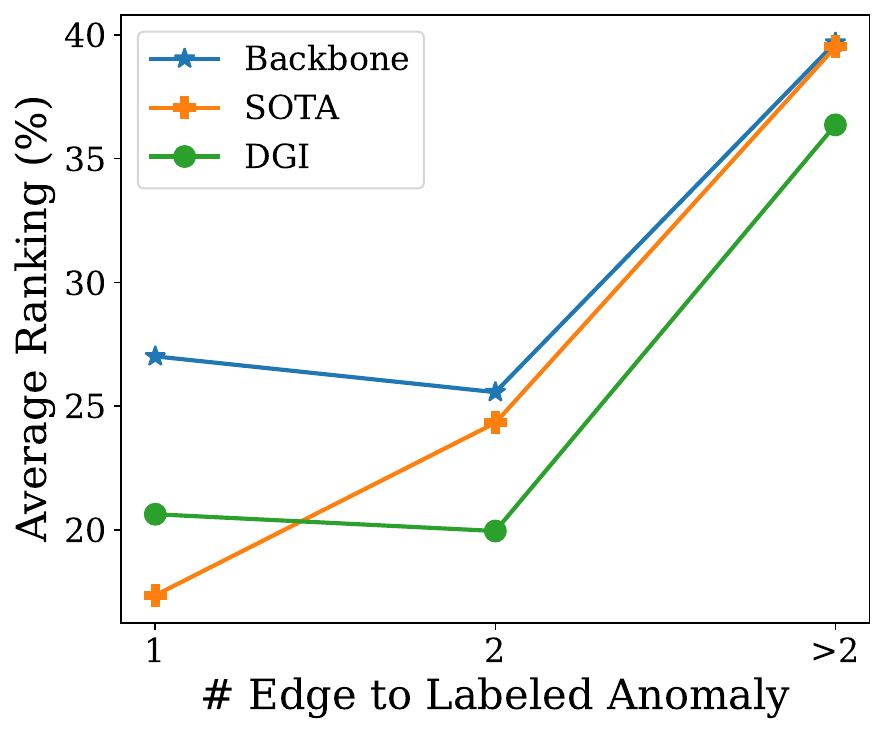}
    \end{minipage}
    }
    \subfigure[T-Social ($R_{2}$=0.40)]{
    \centering
    \begin{minipage}[b]{0.23\textwidth}
    \includegraphics[width=1.0\textwidth]{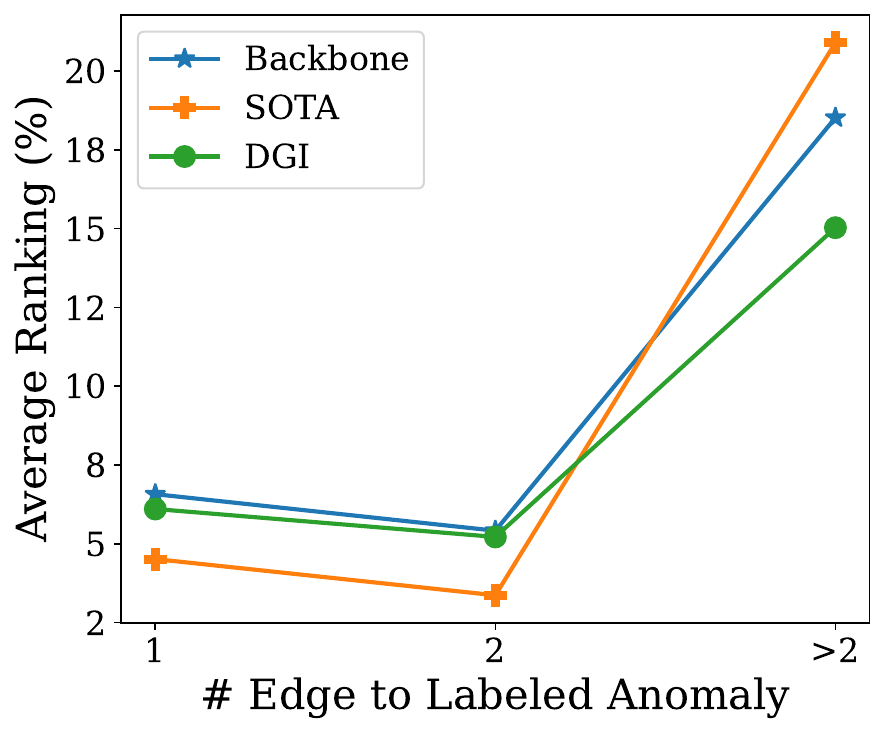}
    \end{minipage}
    }
    \subfigure[Tolokers ($R_{2}$=0.94)]{
    \centering
    \begin{minipage}[b]{0.23\textwidth}
    \includegraphics[width=1.0\textwidth]{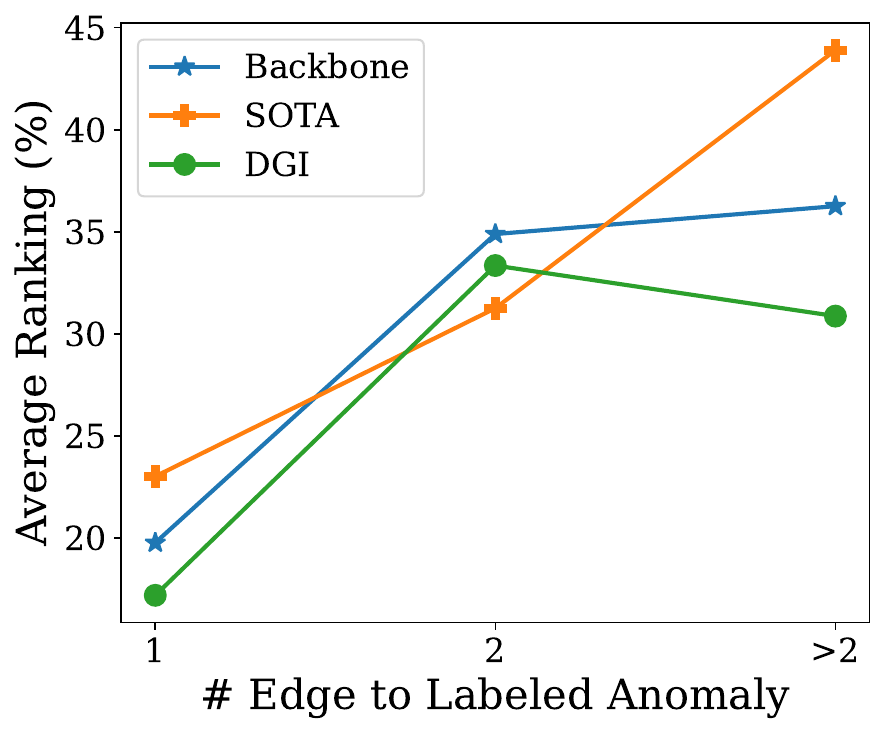}
    \end{minipage}
    }
    \caption{The average ranking for unlabeled anomalies located in the k$^{th}$-hop neighborhood.
    \label{fig:sparse_rank}}
\end{figure*}

Experimental results are presented in Figure~\ref{fig:sparse_rank}.
Notably, DGI consistently exhibits competitive anomaly identification ability across different hop distances. 
Particularly noteworthy are the substantial enhancements observed in detecting anomalies outside the 2-hop neighborhood of labeled anomalies, with performance gains ranging from 5\% to 15\%.
In contrast, while SOTA baselines excel in detecting well-represented 1$^{st}$-hop and 2$^{nd}$-hop anomalies, their performance dramatically deteriorates for anomalies located at least 2 hops away from labeled anomalies.
It is worth noting that, the backbone model, a vanilla GNN, even exhibits a strong detection ability of 1$^{st}$-hop anomalies in YelpChi when compared to its pre-training counterpart.
These observations provide strong empirical support for our hypothesis that the superior performance of pre-training methods is mainly attributed to its improvement in detecting 2-hop unreachable anomalies.
Furthermore, they also signify that end-to-end learning methods tend to disproportionately focus on nearby anomalies while disregarding under-represented anomalies.

\begin{figure}
    \centering
    \includegraphics[width=\linewidth]{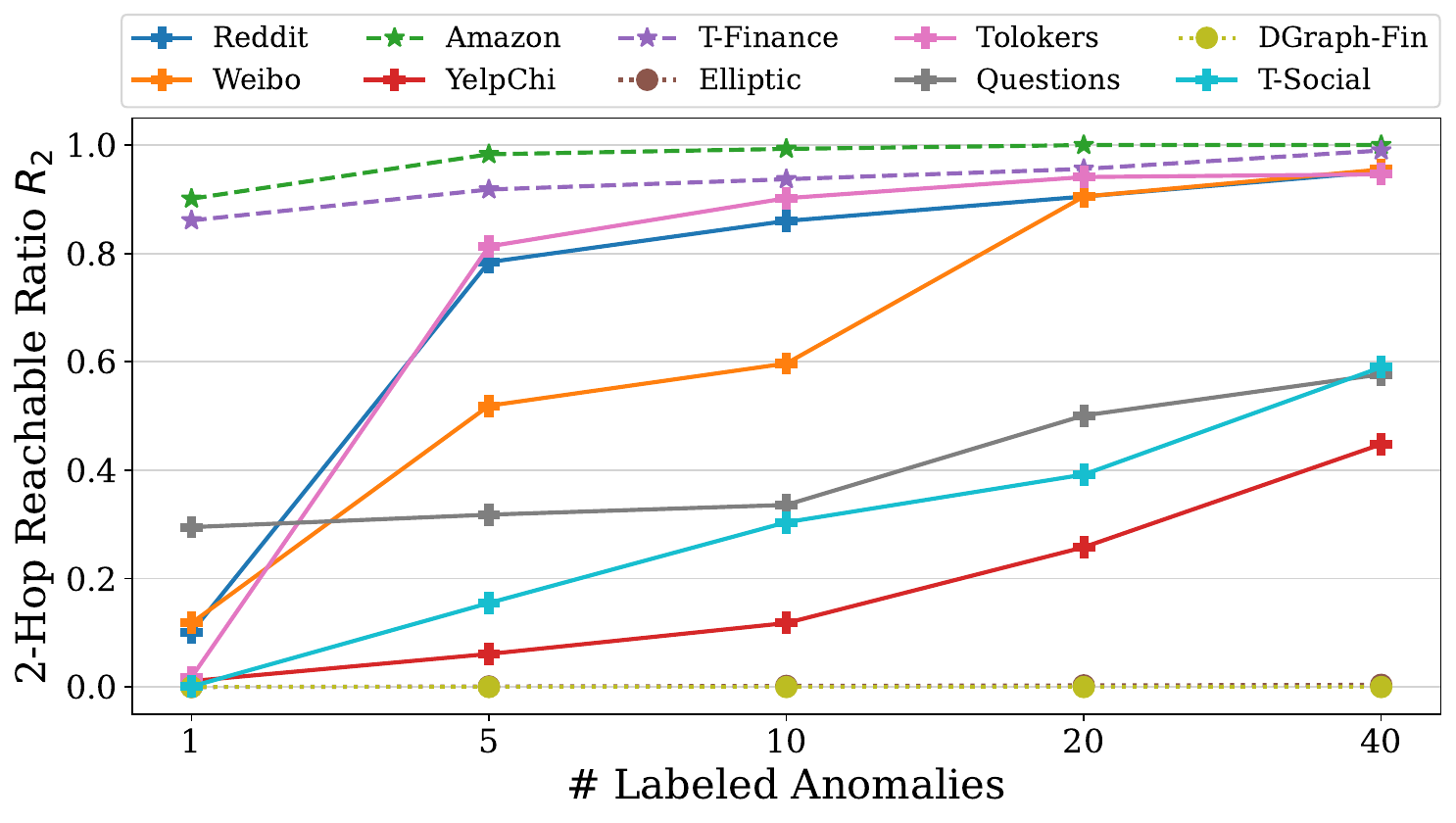}
    \caption{The 2-hop reachable ratio $R_{2}$ with different number of labeled anomalies across all datasets.}
    \label{fig:num_anomaly_reach}
\end{figure}

\begin{figure*}[!h]
    \subfigure[YelpChi]{
    \centering
    \begin{minipage}[b]{0.23\textwidth}
    \includegraphics[width=1.0\textwidth]{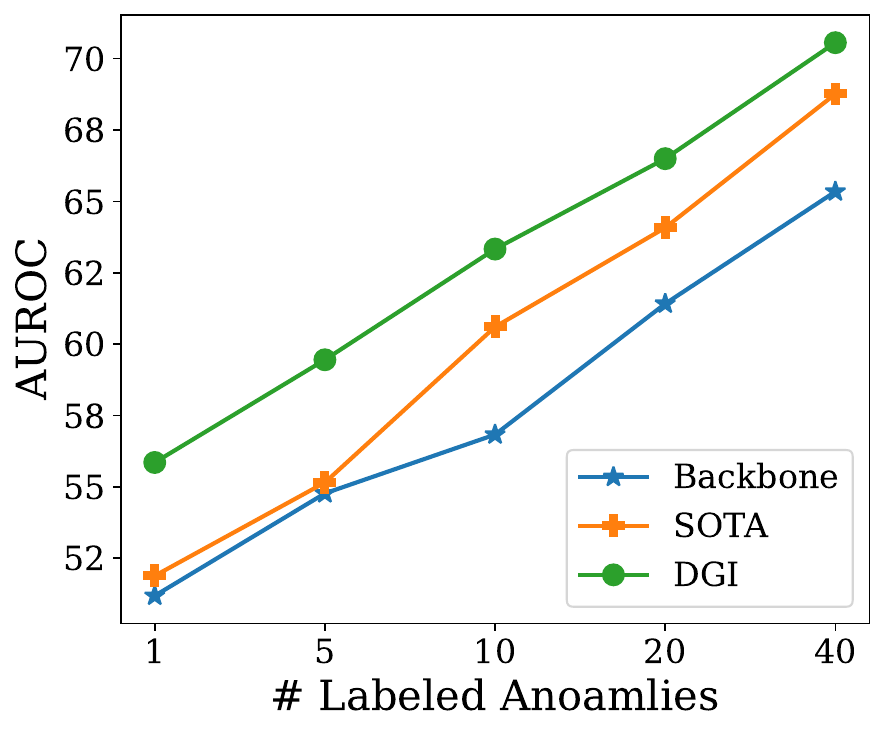}
    \end{minipage}
    }
    \subfigure[Questions]{
    \centering
    \begin{minipage}[b]{0.23\textwidth}
    \includegraphics[width=1.0\textwidth]{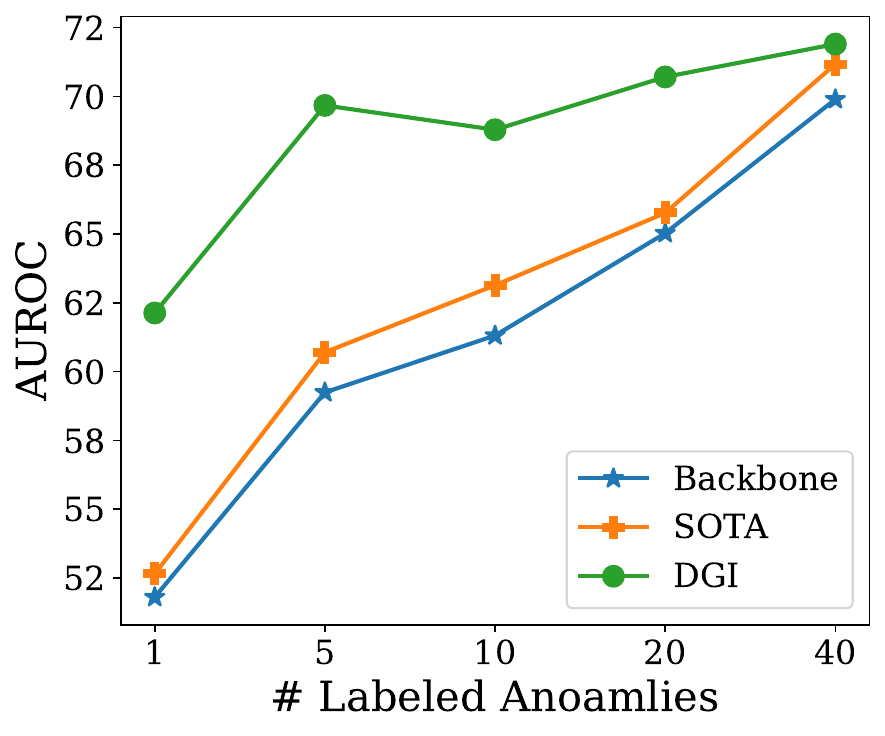}
    \end{minipage}
    }
    \subfigure[T-Social]{
    \centering
    \begin{minipage}[b]{0.23\textwidth}
    \includegraphics[width=1.0\textwidth]{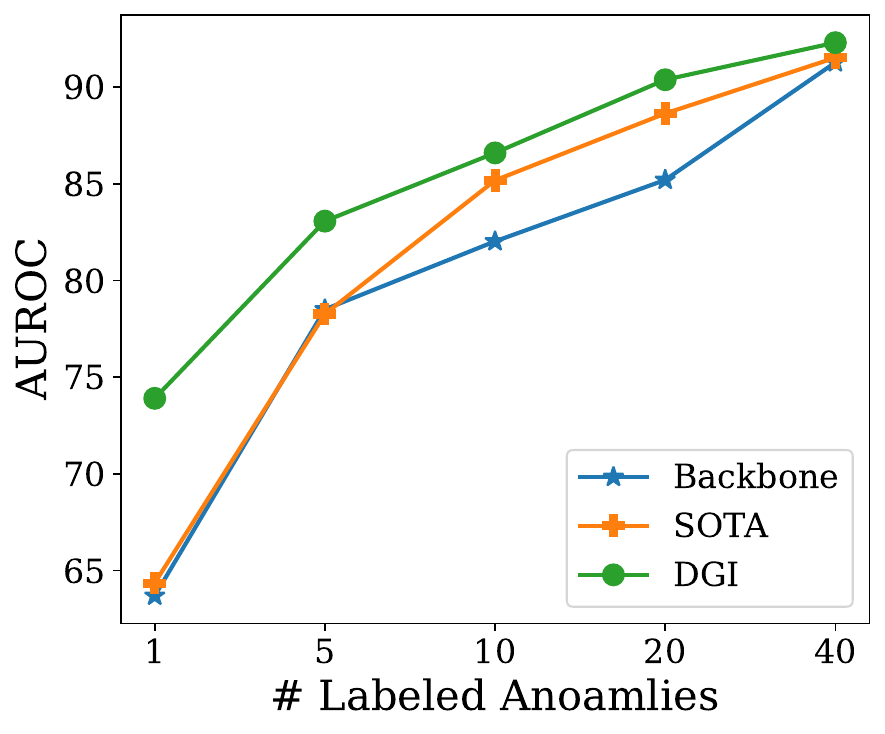}
    \end{minipage}
    }
    \subfigure[Tolokers]{
    \centering
    \begin{minipage}[b]{0.23\textwidth}
    \includegraphics[width=1.0\textwidth]{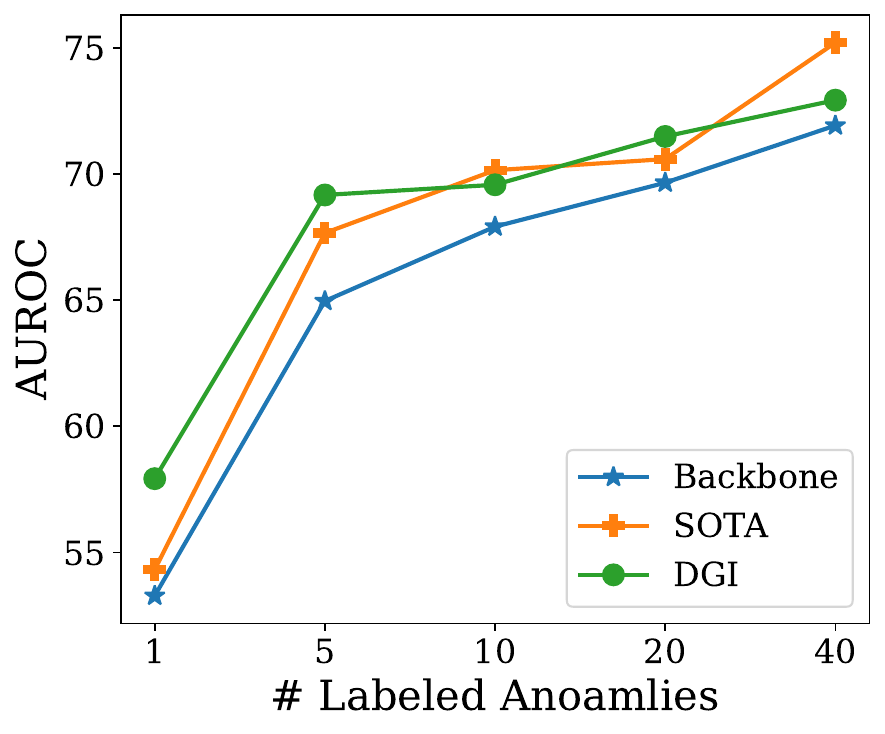}
    \end{minipage}
    }
    \caption{The average AUROC score with the different number of labeled anomalies.
    \label{fig:sparse_num_ano}}
\end{figure*}

\subsection{Key Factors Influencing the Success of Pre-training}

Besides outperforming the SOTA models, it is imperative to acknowledge that the pre-training yields significant advancements for basic backbone GNNs trained in an end-to-end manner.
In essence, the success of pre-training heavily relies on the design of pretext tasks.
We note that the negative sampling in DGI is akin to generating data dissimilar to the original inputs.
This process, we posit, is not merely a contrastive mechanism but rather the creation of `pseudo anomalies'.
In the pre-training stage, the model is tasked with discerning between `normal' and `perturbed' data representations, thereby mirroring the anomaly detection it will later perform.
To validate our hypothesis, we explore how different shuffling ratios for negative sampling in DGI affect the performance of downstream anomaly detection.

Experimental results for all sparse graphs are depicted in Figure~\ref{fig:perturb_ratio}.
Notably, DGI displays a consistent improvement with the increase in perturbation ratios.
For Reddit and YelpChi, there is an upward trend, indicating enhanced detection ability through learning from highly perturbed data during the pre-training stage.
However, Tolokers and Questions exhibit a plateau effect beyond a certain perturbation threshold, suggesting a limit to the benefits of distinguishing perturbed data. 
These findings provide empirical support for our hypothesis that the introduction of `pseudo anomalies' via negative samples during pre-training enhances the model's downstream anomaly detection capabilities.

In addition, drawing from our analyses, we posit that the increased number of labeled anomalies may reduce the relative advantage obtained from the pre-training stage, eventually being overshadowed by direct supervision.
To gain a deeper understanding, we illustrate the variations in the 2-hop reachable ratio $R_{2}$ and the performance of various models under varying quantities of labeled anomalies, as depicted in Figures~\ref{fig:num_anomaly_reach} and ~\ref{fig:sparse_num_ano} respectively.
It can be observed that the advantages conferred by pre-training are most evident when only a single labeled anomaly is available. 
However, as the number of labeled anomalies increases, these advantages over the SOTA models rapidly diminish.
Particularly, as $R_{2}$ approaches 1, the pre-training method exhibits inferior performance compared to SOTA models.
These observations offer empirical validation for our assumption, further indicating that pre-training is well-suited for scenarios with extremely limited supervision.

\vpara{Summary.}
Our study primarily unveils the superior performance of pre-training methods in sparse graphs.
Further probing into this observation, we introduce the k-hop reachable ratio $R_{k}$ as a pivotal metric to evaluate the potential for information propagation.
Through empirical investigation, we establish that the proficiency of pre-training methods predominantly arises from its ability to detect 2-hop unreachable anomalies, with its impact diminishing as $R_{2}$ approaches 1.
Our analyses indicate that pre-training is preferred for GAD in scenarios characterized by sparse graphs and limited supervision, which is indeed commonly encountered in real-world applications.
For additional experimental results, please refer to Appendix~\ref{sec:appendix_exp_setup}.

\begin{table*}[t]
    \centering 
    \caption{Comparison of the AUROC score of each model for the graph-level task.
    The best and runner-up models are highlighted in bold and underlined. 
    Results are averaged across 5 runs.
    The \textcolor{orange}{orange}/\textcolor{blue}{blue} color indicates that pre-training outperforms/underperforms the counterpart.}
    \label{table:graph_auroc}
    \begin{tabular}{c|ccccccc|c}
    \toprule
    Model & DD & IMDB-B & REDDIT-B & PROTEINS & AIDS & NCI1 & Mutagenicity & Average \\
    \midrule
    GCN & 63.79 & 43.68 & 53.44 & 39.98 & 81.39 & 57.85 & 67.64 & 58.25 \\
    GIN & 42.24 & 53.72 & 53.50 & 60.38 & 95.66 & 62.61 & 68.21 & 62.33 \\
    OCGIN & 49.54 & 57.48 & 68.13 & 53.55 & 97.73 & 58.72 & 41.88 & 61.00 \\
    OCGTL & 76.51 & 61.36 & \textbf{86.21} & 62.82 & 97.03 & 59.10 & 41.99 & 69.29 \\
    GLocalKD & 67.66 & \textbf{69.10} & \underline{76.38} & \textbf{68.29} & 95.96 & 58.29 & 57.62 & 70.47 \\
    iGAD & 61.51 & 44.62 & 71.81 & 51.65 & 96.18 & 59.06 & 68.34 & 64.74 \\
    GmapAD & 54.05 & 46.00 & 57.07 & 28.40 & 96.64 & 58.44 & 45.52 & 55.16 \\
    \midrule
    GraphMAE & \underline{78.21} & 59.79 & 73.58 & \underline{64.83} & \textbf{98.40} & \underline{64.07} & \underline{70.18} & \textbf{72.72} \\
    DGI & \textbf{79.22} & \underline{61.95} & 68.35 & 57.93 & \underline{98.07} & \textbf{64.85} & \textbf{70.68} & \underline{71.58} \\
    Imp. Back. & \textcolor{orange}{+ 15.43} & \textcolor{orange}{+ 8.23} & \textcolor{orange}{+ 20.08} & \textcolor{orange}{+ 4.45} & \textcolor{orange}{+ 2.74} & \textcolor{orange}{+ 2.24} & \textcolor{orange}{+ 2.47} & \textcolor{orange}{+ 10.39} \\
    Imp. SOTA  & \textcolor{orange}{+ 2.71} & \textcolor{blue}{- 7.15} & \textcolor{blue}{- 12.63} & \textcolor{blue}{- 3.46} & \textcolor{orange}{+ 0.67} & \textcolor{orange}{+ 2.24} & \textcolor{orange}{+ 2.34} & \textcolor{orange}{+ 2.25}\\
    \bottomrule
    \end{tabular}
\end{table*}

\section{Potential of Pre-Training in Graph-Level Anomaly Detection}

Motivated by the demonstrated successes outlined in prior analyses, we endeavor to explore the potential of pre-training for graph-level anomaly detection. 

\vpara{Graph-level anomaly detection.}
Similar to node-level task, given a collection of attributed graphs $\mathbb{G} = \{ \mathcal{G}_{i} = (\mathcal{V}_{i}, \mathbf{A}_{i}, \mathbf{X}_{i}) \}_{i=1}^{N}$ with the set of partial labels, our objective is to identify whether an individual graph $\mathcal{G}_{i}$ is anomaly or not. 

Compared to node-level anomaly detection, graph-level anomaly detection presents heightened complexities due to anomalies potentially stemming from unique spatial structures and nodal attributes within individual graphs, as well as cross-graph structural and attribute patterns. 
If pre-training can continue to excel in graph-level anomaly detection, it may offer valuable insights for future research directions.
Motivated by our previous findings, we mainly focus on semi-supervised scenarios here.
In the following context, we provide an empirical study of the effectiveness of pre-training models in graph-level anomaly detection.

\vpara{Configurations.}
Following the previous works~\cite{zhao2021using,ma2023towards}, we select 7 public datasets from TUDataset~\cite{Morris+2020} and downsample one specific class, retaining only 10\% of its data samples, to serve as anomalies, as detailed in Table~\ref{table:graph_datasets} in the Appendix~\ref{sec:appendix_exp_setup}.
Specifically, DD~\cite{shervashidze2011weisfeiler} and PROTEINS~\cite{borgwardt2005protein} are designed to identify enzymes within protein structure graphs.
Mutagenicity~\cite{kazius2005derivation}, AIDS~\cite{riesen2008iam}, and NCI1~\cite{shervashidze2011weisfeiler} aim to detect mutagens, HIV activity, and anti-HIV inactivity among molecular networks.
IMDB-BINARY~\cite{cai2018simple} and REDDIT-BINARY~\cite{yanardag2015deep} concentrate on identifying different movie genres and subreddit types within social networks.
The end-to-end learning baselines are GNNs designed for graph-level anomaly detection including OCGIN~\cite{zhao2021using}, OCGTL~\cite{qiu2022Raising}, GLocalKD~\cite{ma2022deep}, iGAD~\cite{zhang2022dual}, GmapAD~\cite{ma2023towards}.
We mainly explore semi-supervised scenarios where the training ratio is set to be 5\% as suggested by~\cite{zhao2021using,CareGNN}.
We report the average performances of 5 runs to ensure robustness.
To maintain consistency, the remaining configurations are identical to the node-level tasks.
Further details, including model configurations and hyperparameter optimization, are provided in Appendix~\ref{sec:appendix_exp_setup}.

\vpara{Results analysis.}
As depicted in Table~\ref{table:graph_auroc} and~\ref{table:graph_auprc}, pre-training models continue to exhibit promising performance, achieving state-of-the-art results when compared to other end-to-end learning baselines across four out of seven datasets. 
Notably, pre-training models consistently outperform the backbone model across all datasets, yielding more significant improvements compared to those observed in node-level tasks, with an absolute increase of 10.39\% on AUROC and 7.69\% on AUPRC. 
Particularly, more pronounced enhancements are observed in protein structure graphs and social networks, which significantly surpass those seen in molecular graphs.
This discrepancy may be attributed to the larger scale and high average node degrees of these two types of graphs, as evidenced in Table~\ref{table:graph_datasets}.
This may suggest that pre-training methods can leverage abundant additional information from the pretext task to aid anomaly detection in such cases.
When compared to GLocalKD, the best GNNs tailored for graph-level anomaly detection, pre-training methods present an absolute increase of 2.25\% on AUROC and 2.56\% on AUPRC.
In summary, even with a trivial GNN backbone, pre-training continues to strive in graph-level anomaly detection under limited supervision.

\section{Conclusion}

In this paper, we conduct a systematic study to explore the crucial problems: \textit{when} and \textit{why} pre-training models excel in GAD, demonstrating through rigorous experiments that pre-training, even with basic GNN architectures, significantly outperforms state-of-the-art models under limited supervision.
Our investigation reveals a critical correlation between pre-training success and graph density,
We further uncover the superiority of pre-training against end-to-end training stems from the enhanced detection of distant, under-represented, unlabeled anomalies that go beyond 2-hop neighborhoods of known anomalies.
Additionally, we show that negative sampling creates 'pseudo anomalies', enhancing detection in downstream tasks, although the benefits decrease with more supervision.
Moreover, we explore and validate the potential of pre-training in the more complex task of graph-level anomaly detection.
Our study provides valuable insights to re-evaluate the role of pre-training in GAD.
\bibliographystyle{ACM-Reference-Format}
\bibliography{main}

\clearpage
\appendix
\section{Additional Experimental Details}\label{sec:appendix_exp_setup}

\vpara{Hyperparamters settings.}
Table~\ref{table:hyperparameters} provides a comprehensive list of search spaces of all hyperparameters.
For all configurations, we retain the model that yields the best AUPRC score on the validation set and report the corresponding test performance.

\begin{table}[h]
    \centering 
    \caption{Default hyperparameters and search space for all models.}
    \label{table:hyperparameters}
    \resizebox{\columnwidth}{!}{
    \begin{tabular}{ccc}
    \toprule
    \textbf{Hyperparameter} & \textbf{Default value} & \textbf{Search Space} \\
    \midrule
    learning rate & 0.005 & [0.01, 0.005, 0.001] \\
    hidden dimension & 32 & [32,64] \\
    layers & 2 & [1,2,3] \\
    activation & ReLU & [ReLU, LeakyReLU, Tanh] \\
    epochs & 200 & [100, 200, ..., 1000] \\
    \bottomrule
    \end{tabular}
    }
\end{table}

\vpara{Implementation specifications.}
To ensure a comprehensive evaluation and maintain fairness across a broad spectrum of models, all baseline models are reproduced in the same environment with PyTorch 1.13, PyTorch Geometric 2.4, and DGL 1.1.

\vpara{Computational hardware.}
Specifically, all our experiments were carried out on a Linux server equipped with an AMD EPYC 7763 64-Core CPU processor, 256GB RAM, and an NVIDIA RTX A6000 GPU with 48G memory.

\vpara{Additional experimental results.}
In terms of the train-test split in fully-supervised scenarios, please refer to Table~\ref{table:node_datasets_split}.
For the results of AUPRC scores across both settings, please refer to Table~\ref{table:node_semi_auprc} and~\ref{table:node_full_auprc} respectively.
For graph-level anomaly detection, we present the data statistics in Table~\ref{table:graph_datasets} in which the specific class chosen to be downsampled as anomaly is highlighted.
For the results of AUPRC scores for graph-level tasks, please refer to Table~\ref{table:graph_auprc}.

\begin{table}[h]
    \centering 
    \caption{Additional information of graph anomaly detection datasets.}
    \label{table:node_datasets_split}
    \resizebox{\columnwidth}{!}{
    \begin{tabular}{ccr}
    \toprule
    Dataset & Train Ratio & Detailed Feature Description \\
    \midrule
    Reddit & 40\% & LIWC text embedding for posts \\
    Weibo & 40\% & Bag-of-words features from posts \\
    Amazon & 70\% & Hand-crafted user features and statistics \\
    YelpChi & 70\% & Hand-crafted review features and statistics \\
    Tolokers & 50\% &  User profile with task performance statistics \\
    Questions & 50\% & FastText embeddings for user descriptions \\
    T-Finance & 40\% & User profile details such as registration days \\
    Elliptic & 50\% & Timestamps and transaction information \\
    DGraph-Fin & 70\% & Timestamps and user profiles details \\
    T-Social & 40\% & User profile details such as logging activities \\
    \bottomrule
    \end{tabular}
    }
\end{table}

\begin{table*}[t]
    \centering 
    \caption{Comparison of the AUPRC score of each model in semi-supervised setting. 
    The best and runner-up models are highlighted in bold and underlined. 
    Results are averaged across 10 runs.
    The \textcolor{orange}{orange}/\textcolor{blue}{blue} color indicates that pre-training outperforms/underperforms the counterpart.}
    \label{table:node_semi_auprc}
    \resizebox{\textwidth}{!}{
    \begin{tabular}{c|cccccccccc|c}
    \toprule
    Model & Reddit & Weibo & Amazon & YelpChi & T-Finance & Elliptic & Tolokers & Questions & DGraph-Fin & T-Social & Average \\
    \midrule
    GCN & 4.67 & 89.01 & 32.83 & 16.47 & 71.40 & 43.69 & 37.31 & 9.12 & 2.27 & 40.74 & 34.75 \\
    GIN & 4.34 & 67.62 & 66.41 & 26.25 & 44.83 & 40.12 & 31.82 & 6.74 & 2.04 & 7.07 & 29.72 \\
    PCGNN & 3.44 & 69.33 & 81.92 & 25.03 & 58.12 & 40.32 & 33.91 & 6.42 & 2.41 & 8.02 & 32.89 \\
    GAT-sep & 4.63 & 76.52 & 80.93 & 24.43 & 34.22 & 46.82 & 33.63 & 7.42 & \textbf{2.43} & 10.43 & 32.15 \\
    BWGNN & 4.22 & 80.62 & \underline{81.73} & 23.74 & 60.93 & 43.44 & 35.33 & 6.53 & 2.27 & 15.92 & 35.47 \\
    GHRN & 4.24 & 77.03 & 80.72 & 23.81 & 63.44 & 44.21 & 35.91 & 6.51 & 2.33 & 16.23 & 35.44 \\
    RF-Graph & 4.39 & 73.70 & 70.72 & 23.62 & \textbf{81.12} & \textbf{80.53} & 35.81 & \underline{10.14} & 1.98 & \textbf{51.33} & 43.33 \\
    XGB-Graph & 4.09 & 82.91 & \textbf{84.38} & 24.86 & \underline{78.36} & \underline{77.21} & 32.08 & 7.76 & 1.91 & 40.62 & \underline{43.42} \\
    \midrule
    GraphMAE & \underline{5.22} & \underline{94.13} & 75.22 & \textbf{29.49} & 75.14 & 54.17 & \underline{39.33} & 8.94 & 2.24 & 45.11 & 42.90 \\
    DGI & \textbf{5.32} & \textbf{94.04} & 78.51 & \underline{29.02} & 70.83 & 59.14 & \textbf{39.59} & \textbf{11.33} & \underline{2.43} & \underline{45.62} & \textbf{43.58} \\
    Imp. Back. & \textcolor{orange}{+ 0.65} & \textcolor{orange}{+ 5.03} & \textcolor{orange}{+ 12.10} & \textcolor{orange}{+ 3.24} & \textcolor{orange}{+ 3.74} & \textcolor{orange}{+ 15.45} & \textcolor{orange}{+ 2.28} & \textcolor{orange}{+ 2.21} & \textcolor{orange}{+ 0.16} & \textcolor{orange}{+ 4.88} & \textcolor{orange}{+ 8.83} \\
    Imp. SOTA. & \textcolor{orange}{+ 0.65} & \textcolor{orange}{+ 5.03} & \textcolor{blue}{- 5.87} & \textcolor{orange}{+ 3.24} & \textcolor{blue}{- 5.98} & \textcolor{blue}{- 21.98} & \textcolor{orange}{+ 2.28} & \textcolor{orange}{+ 1.19} & 0.00 & \textcolor{blue}{- 5.71} & \textcolor{orange}{+ 0.16} \\
    \bottomrule
    \end{tabular}
    }
\end{table*}
\begin{table*}[t]
    \centering 
    \caption{Comparison of the AUPRC score of each model in fully-supervised setting. 
    The best and runner-up models are highlighted in bold and underlined. 
    Results are averaged across 10 runs.
    The \textcolor{orange}{orange}/\textcolor{blue}{blue} color indicates that pre-training outperforms/underperforms the counterpart.}
    \label{table:node_full_auprc}
    \resizebox{\textwidth}{!}{
    \begin{tabular}{c|cccccccccc|c}
    \toprule
    Model & Reddit & Weibo & Amazon & YelpChi & T-Finance & Elliptic & Tolokers & Questions & DGraph-Fin & T-Social & Average \\
    \midrule
    GCN & 4.63 & 94.64 & 45.65 & 20.88 & 78.22 & 25.37 & 40.57 & 14.06 & \underline{3.80} & 76.35 & 40.41  \\
    GIN & 6.41 & 91.67 & 84.61 & 33.63 & 78.35 & 26.21 & 40.36 & 13.68 & 3.47 & 60.79 & 43.92  \\
    PCGNN &  7.73 & 89.07 & 89.33 & 44.51 & 83.31 & 42.66 & 44.85 & 15.59 & 3.42 & 80.29 & 50.08  \\
    GAT-sep & 7.19 & 93.40 & 84.72 & 45.49 & 84.01 & 26.35 & 46.66 & 17.90 & 3.84 & 33.39 & 44.28 \\
    BWGNN & 8.32 & 94.01 & \underline{91.48} & 61.53 & 89.38 & 29.31 & 49.58 & \textbf{18.57} & \textbf{3.97} & 78.93 & 52.51 \\
    GHRN & 4.66 & 95.27 & 89.52 & 55.42 & 87.60 & 43.90 & 47.45 & \underline{18.31} & 3.80 & 86.78 & 53.27  \\
    RF-Graph & 5.13 & \underline{96.95} & 90.53 & \underline{83.92} & \underline{89.23} & \textbf{78.86} & \underline{52.34} & 14.44 & 2.15 & \textbf{97.63} & \underline{61.12} \\
    XGB-Graph & 5.29 & \textbf{97.06} & \textbf{93.33} & \textbf{91.11} & \textbf{90.12} & \underline{77.78} & \textbf{53.92} & 18.19 & 3.79 & \underline{97.34} & \textbf{62.79} \\
    \midrule
    GraphMAE & \underline{7.27} & 94.13 & 88.26 & 60.83 & 81.61 & 50.66 & 50.67 & 15.41 & 3.48 & 75.32 & 52.76 \\
    DGI & \textbf{9.16} & 96.52 & 88.92 & 58.71 & 79.72 & 59.14 & 50.95 & 15.95 & 3.54 & 79.11 & 54.18 \\
    Imp. Back. & \textcolor{orange}{+ 2.75} & \textcolor{orange}{+ 1.88} & \textcolor{orange}{+ 3.65} & \textcolor{orange}{+ 27.20} & \textcolor{orange}{+ 3.26} & \textcolor{orange}{+ 32.93} & \textcolor{orange}{+ 10.38} & \textcolor{orange}{+ 1.89} & \textcolor{blue}{- 0.26} & \textcolor{orange}{+ 2.76} & \textcolor{orange}{+ 10.26} \\
    Imp. SOTA. & \textcolor{orange}{+ 0.84} & \textcolor{blue}{- 0.54} & \textcolor{blue}{- 5.06} & \textcolor{blue}{- 30.28} & \textcolor{blue}{- 8.51} & \textcolor{blue}{- 19.72} & \textcolor{blue}{- 2.97} & \textcolor{blue}{- 2.62} & \textcolor{blue}{- 0.43} & \textcolor{blue}{- 18.52} & \textcolor{blue}{- 8.61} \\
    \bottomrule
    \end{tabular}
    }
\end{table*}
\begin{table*}[ht]
    \centering 
    \caption{Statistics of graph-level anomaly detection datasets.}
    \label{table:graph_datasets}
    \begin{tabular}{cccccccc}
    \toprule
    Dataset & Domain & Class & \# Graphs & \# Sampled Graphs & Avg. \# Nodes & Avg. \# Edges & Avg. Degree \\
    \midrule
    \multirow{2}{*}{DD} & \multirow{2}{*}{Protein structure} & 0 & 691 & 69 & 355.2 & 1806.6 & 5.04 \\
    & & 1 & 487 & 487 & 183.7 & 898.8 & 4.88 \\
    \midrule
    \multirow{2}{*}{IMDB-B} & \multirow{2}{*}{Social networks} & 0 & 500 & 50 & 20.1 & 193.5 & 9.1 \\
    & & 1 & 500 & 500 & 19.4 & 192.5 & 8.6 \\
    \midrule
    \multirow{2}{*}{REDDIT-B} & \multirow{2}{*}{Social networks} & 0 & 1000 & 100 & 641.3 & 1471.9 & 2.44 \\
    & & 1 & 1000 & 1000 & 218.0 & 519.1 & 2.27 \\
    \midrule
    \multirow{2}{*}{PROTEINS} & \multirow{2}{*}{Protein structure} & 0 & 663 & 66 & 50 & 188.1 & 3.79 \\
    & & 1 & 450 & 450 & 22.9 & 83.1 & 3.64 \\
    \midrule
    \multirow{2}{*}{AIDS} & \multirow{2}{*}{Molecular} & 0 & 400 & 40 & 37.6 & 80.5 & 2.13 \\
    & & 1 & 1600 & 1600 & 10.2 & 20.3 & 1.98 \\
    \midrule
    \multirow{2}{*}{NCI1} & \multirow{2}{*}{Molecular} & 0 & 2053 & 205 & 25.65 & 55.3 & 2.15 \\
    & & 1 & 2057 & 2057 & 34.07 & 73.9 & 2.17 \\
    \midrule
    \multirow{2}{*}{Mutagenicity} & \multirow{2}{*}{Molecular} & 0 & 2401 & 240 & 29.3 & 60.5 & 2.1 \\
    & & 1 & 1936 & 1936 & 31.4 & 62.7 & 2.0 \\
    \bottomrule
    \end{tabular}
\end{table*}
\begin{table*}[t]
    \centering 
    \caption{Comparison of the AUPRC score of each model for the graph-level task.
    The best and runner-up models are highlighted in bold and underlined. 
    Results are averaged across 5 runs.
    The \textcolor{orange}{orange}/\textcolor{blue}{blue} color indicates that pre-training outperforms/underperforms the counterpart.}
    \label{table:graph_auprc}
    \begin{tabular}{c|ccccccc|c}
    \toprule
    Model & DD & IMDB-B & REDDIT-B & PROTEINS & AIDS & NCI1 & Mutagenicity & Average \\
    \midrule
    GCN & 18.05 & 12.05 & 17.34 & 13.59 & 48.02 & 11.66 & 22.11 & 20.40 \\
    GIN & 13.05 & 16.02 & 28.81 & 21.13 & 86.74 & 13.66 & 24.1 & 29.07 \\
    OCGIN & 19.24 & 13.07 & 34.17 & 24.29 & \underline{95.62} & 11.10 & 9.07 & 29.51 \\
    OCGTL & 29.94 & 13.60 & \textbf{53.30} & \textbf{27.59} & 95.51 & 10.52 & 8.91 & 34.20 \\
    GLocalKD & 22.09 & \textbf{22.70} & 28.52 & 23.31 & 44.36 & 11.29 & 12.25 & 23.50 \\
    iGAD & 22.01 & 11.32 & \underline{47.48} & 19.36 & 86.74 & 14.00 & 25.15 & 32.29 \\
    GmapAD & 12.86 & 8.80 & 46.92 & 8.51 & 70.42 & 12.37 & 12.16 & 24.58 \\
    \midrule
    GraphMAE & \underline{31.17} & 17.95 & 42.25 & \underline{27.28} & \textbf{97.18} & \textbf{16.14} & \underline{25.36} & \textbf{36.76} \\
    DGI & \textbf{33.30} & \underline{19.27} & 41.40 & 25.41 & 93.82 & \underline{15.19} & \textbf{27.10} & \underline{36.50} \\
    Imp. Back. & \textcolor{orange}{+ 15.25} & \textcolor{orange}{+ 3.25} & \textcolor{orange}{+ 13.44} & \textcolor{orange}{+ 6.15} & \textcolor{orange}{+ 10.44} & \textcolor{orange}{+ 2.48} & \textcolor{orange}{+ 3.00} & \textcolor{orange}{+ 7.69} \\
    Imp. SOTA. & \textcolor{orange}{+ 3.36} & \textcolor{blue}{- 3.43} & \textcolor{blue}{- 11.05} & \textcolor{blue}{- 0.31} & \textcolor{orange}{+ 1.56} & \textcolor{orange}{+ 2.14} & \textcolor{orange}{+ 1.95} & \textcolor{orange}{+ 2.56} \\
    \bottomrule
    \end{tabular}
\end{table*}

\end{document}